\pdfoutput=1

\documentclass[11pt]{article}

\usepackage[preprint]{acl}

\usepackage{times}
\usepackage{latexsym}

\usepackage[T1]{fontenc}

\usepackage[utf8]{inputenc}

\usepackage{microtype}

\usepackage{inconsolata}

\usepackage{graphicx}
\usepackage{color}
\usepackage{latexsym}
\usepackage{amsmath, amssymb, bm, amsfonts, latexsym, mathtools}
\usepackage{url}
\usepackage {diagbox}
\usepackage[ruled, linesnumbered, titlenumbered]{algorithm2e}
\usepackage{bbm}
\usepackage{multirow}
\usepackage {booktabs}
\usepackage{colortbl}
\usepackage{subcaption}
\usepackage{color}
\usepackage{latexsym}
\usepackage{makecell}
\usepackage{amsmath,amssymb,bm}
\usepackage{url}
\usepackage {diagbox}
\usepackage[ruled, linesnumbered, titlenumbered]{algorithm2e}
\usepackage{bbm}
\usepackage{multirow}
\usepackage {booktabs}
\usepackage{pifont}
\usepackage{siunitx}
\usepackage{array}
\usepackage{colortbl}
\usepackage{ulem}
\usepackage{listings}

\definecolor{red}{HTML}{FFD6D6}
\newcommand{\red}[1]{\cellcolor{red}{#1}}

\newcommand{\sig}[1]{\cellcolor{red}{\textcolor{purple}{#1}}}

\title{IMPARA-GED: Grammatical Error Detection is Boosting \\Reference-free Grammatical Error Quality Estimator}

\author{Yusuke Sakai\textsuperscript{\dag}, \
    Takumi Goto\textsuperscript{\dag}, \
  Taro Watanabe \\
  Nara Institute of Science and Technology (NAIST), Japan \\
  \texttt{\{sakai.yusuke.sr9, goto.takumi.gv7, taro\}@is.naist.jp} \\
  \small \dag Equal Contribution
}

\begin{document}
\maketitle
\begin{abstract}
We propose IMPARA-GED, a novel reference-free automatic grammatical error correction (GEC) evaluation method with grammatical error detection (GED) capabilities.
We focus on the quality estimator of IMPARA, an existing automatic GEC evaluation method, and construct that of IMPARA-GED using a pre-trained language model with enhanced GED capabilities. Experimental results on SEEDA, a meta-evaluation dataset for automatic GEC evaluation methods, demonstrate that IMPARA-GED achieves the highest correlation with human sentence-level evaluations.
\end{abstract}

\section{Introduction}

Grammatical error correction (GEC) is the task of automatically correcting grammatical or superficial errors in input sentences.
While GEC systems should manually assess the quality of their corrections, human evaluation for a wide range of arbitrary inputs is strenuous, making it impractical.
Therefore, it is necessary to establish automatic evaluation methods for GEC that correlate highly with human evaluation.
The automatic GEC evaluation methods can be categorized into reference-based evaluation methods~\cite{bryant-etal-2017-automatic, gotou-etal-2020-taking, koyama-etal-2024-n-gram} and reference-free evaluation methods~\cite{yoshimura-etal-2020-reference, islam-magnani-2021-end, maeda-etal-2022-impara}.

Reference-based evaluation methods measure the closeness between the outputs of GEC systems and human-written references. However, since incorrect sentences can be corrected in multiple ways, accurate evaluation requires multiple reference sentences. Yet, constructing comprehensive human references is impractical, and low-coverage reference sets often deteriorate evaluation reliability~\cite{choshen-abend-2018-automatic, choshen-abend-2018-inherent}. Therefore, reference-free evaluation methods, which rely only on input sentences and system outputs, have the potential to overcome these limitations.

Most reference-free evaluation methods employ pre-trained language models (PLMs).
For instance, Scribendi Score~\cite{islam-magnani-2021-end} uses the perplexity of a PLM to compute evaluation scores for corrected outputs. 
SOME~\cite{yoshimura-etal-2020-reference} trains PLMs separately on human assessment scores for fluency, grammaticality, and meaning preservation.
IMPARA~\cite{maeda-etal-2022-impara} combines a similarity estimator between inputs and system outputs with a quality estimator for system outputs, which rely on PLMs.
The quality estimator is trained without requiring human-annotated evaluation results, using only parallel data of erroneous and corrected sentences constructed for GEC systems.
However, although the quality estimator is trained on a vanilla PLM, its pre-trained knowledge alone is insufficient to capture grammatical errors accurately.

In this paper, we propose IMPARA-GED, a novel reference-free automatic evaluation method for GEC.
Inspired by the insight that additional training on the Grammatical Error Detection (GED) task enhances GEC systems~\cite{yuan-etal-2021-multi, kaneko-etal-2020-encoder}, the quality estimator of IMPARA-GED is constructed by first fine-tuning a PLM on the GED task and then applying quality estimator construction method of IMPARA. Moreover, we remove the similarity estimator used in IMPARA, as empirical observations indicate that it fails to effectively capture grammatical errors in the outputs of modern GEC systems.
Our experimental results on SEEDA, a meta-evaluation benchmark~\cite{kobayashi-etal-2024-revisiting}, show that IMPARA-GED achieves the highest correlation with human sentence-level evaluations.

\section{Background: IMPARA}

IMPARA consists of two models: a \textbf{quality estimator} (QE), which assesses the quality of GEC systems outputs, and a \textbf{similarity estimator} (SE), which evaluates the semantic preservation between inputs and system outputs.
While the similarity estimator utilizes a vanilla PLM, the quality estimator is constructed by fine-tuning an encoder-based PLM such as BERT~\cite{devlin-etal-2019-bert}.

\paragraph{Construction of Quality Estimator.}
\label{sec:create-impara-qe}
The quality estimator is constructed by learning the pairwise quality ranking order $(S_+, S_-)$, where $S_+$ has higher quality and $S_-$ has lower quality.
These pairs can be automatically generated from parallel data of incorrect and correct sentences constructed for GEC systems. Specifically, an edit set for transforming an incorrect sentence into a correct sentence is extracted, and the impact of each edit is quantified based on the degree of semantic change when the edit is removed from the correct sentence.
Two different subsets of the edit set are then sampled and applied to the incorrect sentence, generating two different corrected sentences. Since the impact of each edit is already computed, the total impact can be calculated for each subset as a quality.
We regard the higher-quality correction as $S_+$ and the lower-quality correction as $S_-$, forming the training data $\mathcal{T}$.
Finally, the quality estimator $R$ is trained using the training data $\mathcal{T}$ by minimizing the loss function $\mathcal{L}^{\text{QE}}$ as follows:
\begin{equation}
    \label{eq:1}
    \mathcal{L}^{\text{QE}} = \frac{1}{|\mathcal{T}|}\sum_{(S_+, S_-) \in \mathcal{T}} \sigma(R(S_-) - R(S_+)),
\end{equation}
where $\sigma(\cdot)$ is a sigmoid function.
The quality estimator $R$ linearly transforms the final layer's first token embedding representation of the incorrect sentence into a real-valued score.

\paragraph{Scoring in IMPARA.}
IMPARA calculates the evaluation score $S(I, O) \in [0, 1]$ for a pair of an input sentence $I$ and its corrected output $O$ from a GEC system using the quality estimator $R$ and the similarity estimator $\text{sim}(\cdot)$ as follows:
\begin{equation}
    \label{eq:2}
    S(I, O) = \left\{
\begin{array}{ll}
\sigma(R(O))& \text{sim}(I, O; \text{PLM}) > \theta \\
0 & \text{otherwise}
\end{array}
\right.,
\end{equation}
where $\theta$ is the threshold of the similarity.
This threshold is used to filter out sentences that are irrelevant to the input sentences but receive high quality estimation scores.

\section{Proposed Method: IMPARA-GED}
We propose IMPARA-GED by removing the similarity estimator from IMPARA and incorporating additional training on the GED task before quality estimator construction.

\subsection{Rethinking of the Similarity Estimator}
\label{sec:similarity}

\begin{table}[t]
\centering
\small
\rowcolors{2}{gray!10}{white}
\setlength{\tabcolsep}{1.25pt}
  \begin{tabular}{@{}lcccccccc@{}} \toprule
  \rowcolor{white}
  & \multicolumn{4}{@{}c@{}}{System-level} & \multicolumn{4}{@{}c@{}}{Sentence-level} \\ \cmidrule(lr){2-5} \cmidrule(rl){6-9}
  & \multicolumn{2}{@{}c@{}}{\texttt{SEEDA-S}} & \multicolumn{2}{@{}c@{}}{\texttt{SEEDA-E}} & \multicolumn{2}{@{}c@{}}{\texttt{SEEDA-S}} & \multicolumn{2}{@{}c@{}}{\texttt{SEEDA-E}} \\ \cmidrule(rl){2-3} \cmidrule(lr){4-5} \cmidrule(rl){6-7} \cmidrule(lr){8-9}
\rowcolor{white}
  IMPARA-SE   & $r$ & $\rho$ & $r$ & $\rho$ & Acc. & $\tau$ & Acc. & $\tau$ \\ \midrule
QE only  & .916 & .902 & .902 & .965 & .753 & .506 & .752 & .504 \\ 
\textbf{$^{\dag}$BERT$_{\text{Base}}$} & .916 & .902 & .902 & .965 & .753 & .506 & .752 & .504 \\ \midrule
BERT$_{\text{Large}}$  & .889 & .867 & .909 & .916 & .731 & .463 & .737 & .474 \\ 
BERT$_{\text{Base-uncased}}$  & .922 & .909 & .903 & .944 & .746 & .493 & .745 & .491 \\
BERT$_{\text{Large-uncased}}$ & .902 & .895 & .904 & .951 & .738 & .476 & .743 & .487 \\ 
ELECTRA$_{\text{Base}}$  & .920 & .902 & .904 & .965 & .752 & .505 & .751 & .503 \\ 
\textbf{$^{\dag}$ELECTRA$_{\text{Large}}$}  & .916 & .902 & .902 & .965 & .753 & .506 & .752 & .504 \\ 
DeBERTa-v3$_{\text{Base}}$  & .906 & .916 & .891 & .958 & .750 & .500 & .749 & .498 \\ 
DeBERTa-v3$_{\text{Large}}$  & .915 & .916 & .900 & .958 & .749 & .498 & .749 & .499 \\ 
\textbf{$^{\dag}$ModernBERT$_{\text{Base}}$}  & .916 & .902 & .902 & .965 & .753 & .506 & .752 & .504 \\ 
ModernBERT$_{\text{Large}}$  & .917 & .903 & .903 & .965 & .753 & .505 & .752 & .503 \\ 
\bottomrule
  \end{tabular}
  \caption{The score differences on the SEEDA arise from variations in the PLMs used as the SE. The QE is fixed using the reproduced IMPARA QE weights. Employing BERT$_{\text{Base}}$ as the SE makes it equivalent to the IMPARA. The threshold $\theta$ is 0.9, as in IMPARA. PLMs marked with $\dag$ produced similarity scores above the threshold for all instances in SEEDA, rendering the SE meaningless.}
  \label{tab:impara-se}
\end{table}

We observed that, in some cases, filtering by the similarity threshold did not work properly. This is due to PLMs struggling to effectively capture grammatical errors.
Table~\ref{tab:impara-se} shows the meta-evaluation results of IMPARA using SEEDA~\cite{kobayashi-etal-2024-revisiting} (detailed in \S\ref{sec:setting}), where various PLMs are employed as the similarity estimator. These results indicate that the choice of PLM affects performance, and the similarity estimator either does not work or negatively impacts the results.

This reason is that similarity estimation with a vanilla PLM results in incorrect filtering.
For instance, the sentence pair \textit{``I think the family will stay mentally \uline{healty} \uuline{as it is ,} without having \uline{emtional} stress.''} and \textit{``I think the family will stay mentally \uline{healthy} without having \uline{emotional} stress.''} is assigned a similarity score of 0.787 by BERT$_{\text{Large-uncased}}$.
Here, the corrected sentence is simply a revision of the errors in the incorrect sentence, meaning it should not be filtered by the similarity threshold. However, with IMPARA’s default threshold of 0.9, it is incorrectly filtered.
Moreover, removing the correction of ``healthy'' increases the similarity score to 0.926, suggesting that BERT$_{\text{Large-uncased}}$ struggles to understand the semantic impact of spelling errors.
In contrast, there are cases where incorrect corrections that should be filtered are instead accepted. For instance, the pair \textit{``I \uline{like} cats.''} and \textit{``I \uline{dislike} cats.''} is assigned a high similarity score of 0.980 with BERT$_{\text{Base}}$. 
Since negation is not a valid correction in GEC, this correction should be filtered. However, due to the high similarity, it is mistakenly accepted.

These observations suggest that the similarity estimator fails in its intended role.
Furthermore, in the outputs of modern GEC systems included in the SEEDA dataset, corrections that significantly deviate from the original erroneous sentence are rarely encountered in practice.
This issue of adversarial corrections is not unique to IMPARA but is a general problem observed in other evaluation metrics, such as SOME~\cite{islam-magnani-2021-end}. Given this, filtering adversarial corrections should be explored as a separate research direction. 
Therefore, we focus on only the quality estimation performance of IMPARA and eliminate the similarity threshold as follows:
\begin{equation}
    \label{eq:3}
    S(I, O) = \sigma(R(O)).
\end{equation}
\subsection{Additional training on the GED task}

The sentence pairs used for constructing IMPARA’s quality estimator are created based on the impact of each correction. However, as discussed in \S \ref{sec:similarity}, a vanilla PLM may not sufficiently capture errors.
To address this, IMPARA-GED introduces additional training on the GED task to build a quality estimator that more accurately captures token-level error information. Then, IMPARA’s training method is applied to construct the final quality estimator.

Following~\citet{yuan-etal-2021-multi}, the GED model classifies errors at the token level, using four variations of error labels: (1) \textbf{2-class} setting that binarizes tokens as correct or incorrect, (2) \textbf{4-class} setting that categorizes tokens into correct, insertion, deletion, and substitution, (3) \textbf{25-class} setting based on POS categories as defined by ERRANT~\cite{bryant-etal-2017-automatic}, and (4) \textbf{55-class} setting that combines these classifications.
These token-level labels are automatically assigned based on existing parallel data of incorrect and corrected sentences and the alignment results from ERRANT.

Formally, given an erroneous sentence $\boldsymbol{x} = [x_1, x_2, \dots, x_N]$ consisting of $N$ tokens and its corresponding error labels $\boldsymbol{t} = [t_1, t_2, \dots, t_N]$, the model is trained by minimizing the loss function:%
\begin{equation}\label{eq:train-ged}
    \mathcal{L}^{\text{GED}}(\boldsymbol{x}, \boldsymbol{t}) = -\frac{1}{N}\sum_{i=1}^{N} \log p (t_i | x_i, \boldsymbol{x}).
\end{equation}
Next, IMPARA-GED fine-tunes the GED model for IMPARA’s quality estimator following Equation~\ref{eq:1} and performs inference according to Equation~\ref{eq:3}.
The impact calculation is also done using the same GED model.
Note that, instead of embedding the first token as in Equation~\ref{eq:2}, mean pooling over all token embeddings is applied to make more effective use of token-level error detection information.

\section{Experimental Settings}
\label{sec:setting}

\paragraph{Construction of IMPARA-GED.}
We used CoNLL-2013~\cite{ng-etal-2013-conll} and FCE~\cite{yannakoudakis-etal-2011-new} for model construction.
CoNLL-2013 was split into train, dev, and devtest sets in an 8:1:1 ratio, while FCE was used with its predefined splits.
First, we construct the GED model following the settings of~\citet{yuan-etal-2021-multi}. The PLM is trained for five epochs using a combined train set of FCE and CoNLL-2013, and the checkpoint that achieves the highest score on the FCE dev set is selected as the final GED model.
Next, the GED model is used to build the quality estimator following the procedure described in \S~\ref{sec:create-impara-qe} and the settings of ~\citet{maeda-etal-2022-impara}, using the CoNLL-2013 train set. The GED model is then trained for ten epochs following Equation~\ref{eq:1}, and the checkpoint achieving the best performance on the CoNLL-2013 dev set is selected.
We trained the model using five different random seeds, and the one that performs best on the CoNLL-2013 devtest set is selected as the final model.
We report results for all combinations of the following label granularities: 2-class, 4-class, 25-class, and 55-class, and the following PLMs: BERT$_{\text{Base}}$~\cite{devlin-etal-2019-bert},
DeBERTa-v3$_{\text{Large}}$~\cite{he2023debertav}, and ModernBERT$_{\text{Large}}$~\cite{warner2024smarterbetterfasterlonger}.

\paragraph{Evaluations.}
We conduct meta-evaluations using SEEDA~\cite{kobayashi-etal-2024-revisiting}, using both edit-level annotations (\texttt{SEEDA-E}) and sentence-level one (\texttt{SEEDA-S}).
We follow the TrueSkill-based system ranking and the \texttt{Base} system setting, which includes outputs from 12 modern GEC systems.
We report Pearson’s correlation coefficient $r$ and Spearman’s rank correlation coefficient $\rho$ as correlation metrics for system-level evaluation and Accuracy (Acc.) and Kendall’s rank correlation coefficient $\tau$ for sentence-level evaluation.
As baselines, we include reference-based evaluation methods: ERRANT~\cite{felice-etal-2016-automatic, bryant-etal-2017-automatic}, PT-ERRANT~\cite{gong-etal-2022-revisiting}, GREEN~\cite{koyama-etal-2024-n-gram}, and GLEU~\cite{napoles-etal-2015-ground}.
For reference-free evaluation methods, we report Scribendi Score~\cite{islam-magnani-2021-end}, SOME~\cite{yoshimura-etal-2020-reference}, and the original IMPARA~\cite{maeda-etal-2022-impara}.
Additionally, we include GPT-4-S~\cite{kobayashi-etal-2024-large} and its three derivative systems as large language model-based GEC evaluation methods.
Our implementation uses \texttt{gec-metrics}\footnote{\url{https://github.com/gotutiyan/gec-metrics}}~\cite{goto2025gecmetrics} with each system’s default settings.
For GPT-4-S, we cite the reported values from~\citet{kobayashi-etal-2024-large}.
Following~\citet{goto2025}, all system evaluations are conducted using TrueSkill~\cite{NIPS2006_f44ee263}, aligning with the aggregation method used in human evaluation.
We also conducted significance testing following~\citet{yoshimura-etal-2020-reference}, using Williams significance tests~\cite{graham-baldwin-2014-testing} for system-level evaluation and bootstrap resampling~\cite{graham-etal-2014-randomized} for sentence-level evaluation.

\section{Experimental Results and Discussions}

\begin{table}[!t]
\centering
\small
\rowcolors{2}{gray!10}{white}
\setlength{\tabcolsep}{1.4pt}
  \begin{tabular}{@{}lcccccccc@{}} \toprule
  & \multicolumn{4}{@{}c@{}}{System-level} & \multicolumn{4}{@{}c@{}}{Sentence-level} \\ \cmidrule(lr){2-5} \cmidrule(rl){6-9}
  & \multicolumn{2}{@{}c@{}}{\texttt{SEEDA-S}} & \multicolumn{2}{@{}c@{}}{\texttt{SEEDA-E}} & \multicolumn{2}{@{}c@{}}{\texttt{SEEDA-S}} & \multicolumn{2}{@{}c@{}}{\texttt{SEEDA-E}} \\ \cmidrule(rl){2-3} \cmidrule(lr){4-5} \cmidrule(rl){6-7} \cmidrule(lr){8-9}
\rowcolor{white}
 Methods  & $r$ & $\rho$ & $r$ & $\rho$ & Acc. & $\tau$ & Acc. & $\tau$ \\ \midrule

ERRANT  & .763 & .706 & .881 & .895 & .594 & .189 & .608 & .217 \\ 
PT-ERRANT  & .870 & .797 & .924 & .951 & .582 & .165 & .592 & .184 \\ 
GREEN   & .855 & .846 & .912 & \uline{.965} & .600 & .199 & .574 & .148 \\
GLEU    & .863 & .846 & .909 & \uline{.965} & .672 & .343 & .673 & .347 \\
Scribendi  & .674 & .762 & .837 & .888 & .660 & .320 & .672 & .345 \\
SOME       & .932 & .881 & .893 & .944 & .778 & .555 & .766 & .532 \\
IMPARA     & .939 & .923 & .901 & .944 & .753 & .506 & .752 & .504 \\
\midrule  
GPT-4-S    & .887 & .860 & .960 & .958 & .784 & .567 & .798 & .595 \\
\;\;+$_{\text{Grammaticality}}$ & .888 & .867 & \uline{.961} & .937 & .796 & .592 & .807 & .615 \\
\;\;+$_{\text{Fluency}}$ & .913 & .874 & \textbf{.974} & \textbf{.979} & \uline{.819} & \uline{.637} & \textbf{.831} & \textbf{.662} \\
\;\;+$_{\text{ Meaning Preservation}}$ & .958 & .881 & .911 & .960 & .810 & .620 & \uline{.813} & \uline{.626} \\
\midrule  
BERT$_{\text{Base}}$ & .915       & .895       & .875       & .930 & .756       & .512       & .754       & .508       \\ 
\;\;\;\;+2-class     & \red{.916} & \red{.909} & .850       & .902 & \red{.773} & \red{.545} & \red{.763} & \red{.527} \\
\;\;\;\;+4-class     & .908       & \red{.902} & .859       & .923 & \sig{.787} & \sig{.574} & \sig{.774} & \sig{.548} \\
\;\;+25-class        & \red{.925} & \red{.902} & \red{.875} & .923 & \red{.771} & \red{.543} & .752       & .503       \\
\;\;+55-class        & .900       & \red{.902} & .842       & .923 & \red{.763} & \red{.526} & .750       & .499       \\
\midrule
DeBERTa-v3$_{\text{Large}}$ & .960 & \uline{.937} & .912       & .944       & .784       & .568       & .779       & .558       \\ 
\;\;\;\;+2-class            & .951 & .923         & .895       & .916       & \sig{.797} & \sig{.593} & \sig{.784} & \sig{.568} \\
\;\;\;\;+4-class            & .939 & .895         & .899       & .916       & \sig{.793} & \sig{.585} & .772       & .544       \\
\;\;+25-class               & .945 & .930         & .906       & .930       & \sig{.801} & \sig{.602} & \sig{.786} & \sig{.573} \\
\;\;+55-class               & .955 & .930         & \red{.913} & \red{.958} & .782       & .564       & .763       & .527       \\
\midrule
ModernBERT$_{\text{Large}}$ & .949                & .909                & .912       & .937       & .767                & .533                & .749       & .497       \\ 
\;\;\;\;+2-class            & \sig{\uline{.971}}  & \sig{.930}          & \red{.919} & .930       & \sig{\textbf{.829}} & \sig{\textbf{.658}} & \sig{.797} & \sig{.594} \\
\;\;\;\;+4-class            & \sig{.964}          & \red{.916}          & \red{.926} & .923       & \sig{.812}          & \sig{.624}          & \sig{.794} & \sig{.588} \\
\;\;+25-class               & \sig{\textbf{.972}} & \sig{\uline{.937}}  & \red{.933} & \red{.944} & \red{.801}          & \red{.603}          & \red{.783} & \red{.567} \\
\;\;+55-class               & \sig{.965}          & \sig{\textbf{.951}} & .910       & .909       & .749                & .498                & .741       & .483 \\
\bottomrule
  \end{tabular}
    \caption{Meta-evaluation results on SEEDA. Each IMPARA-GED variant is identified by the name of the PLM used for training. The first value in each row shows the result without additional GED training, and the following rows show the results after additional training on the GED task using each class label. \textbf{Bold} scores indicate the best performance, \underline{underline} indicates the second-best, \colorbox{red}{red cells} indicate improvements owing to GED task training, and \colorbox{red}{\textcolor{purple}{purple values}} marks statistically significant improvements ($p < 0.05$) over the version without GED training.}
  \label{tab:results}
\end{table}

\paragraph{Main results.}
Table~\ref{tab:results} shows the SEEDA evaluation results for each automatic GEC evaluation method.
We observe a general improvement in sentence-level evaluation when additional training on the GED task is applied. Notably, using ModernBERT$_{\text{Large}}$ with the 2-class classification GED task achieves the highest performance among all methods in the sentence-level \texttt{SEEDA-S}.
Additionally, in sentence-level \texttt{SEEDA-E}, the model outperforms all models except GPT4-S.
In system-level \texttt{SEEDA-E}, the benefits of GED training were not consistently observed.
This may be attributed to the fact that system-level correlations with human evaluation are already high (above 0.9), leaving limited room for further improvement. Indeed, global trends reflected in human judgments are already well captured by automatic evaluation metrics.
Therefore, further improvement in correlation would require the ability to capture more subtle and fine-grained evaluations at the sentence level.
However, for sentence-level evaluation, the IMPARA-GED series consistently outperformed the base model in most cases, and the impact of GED training was more pronounced.
Since IMPARA evaluates at the sentence level, this outcome is reasonable, and GED training has proven effective in enhancing sentence-level evaluation.

\paragraph{Number of GED types.}
Increasing the number of error classification types for GED training does not necessarily lead to better performance, as even binary classification was sufficient to improve evaluation results.
We suspect that label reliability may be influencing the results. In the 55-class setting, although the labels carry more information, their reliability tends to decrease. In contrast, binary classification conveys less information per label but offers higher overall reliability. In the context of GEC evaluation, label reliability may be more critical than label informativeness.
This observation aligns with the findings of~\citet{yuan-etal-2021-multi}, and our results with IMPARA-GED support this view: the binary setting performed best, followed by the 4-class setting. Since the goal of this task is GEC evaluation, using more reliable labels appears better suited to this objective.

\paragraph{System-level Window Analysis.}

\begin{figure}[t]
    \centering
    \includegraphics[width=\linewidth]{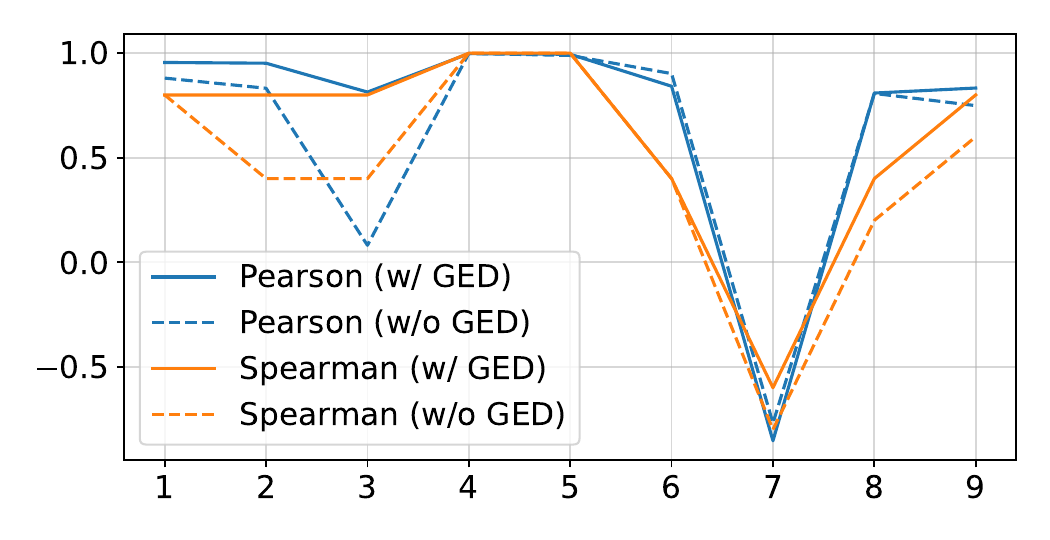}
    \caption{Results of the window analysis with a window size of $4$. The x-axis represents the starting rank in the human rankings; for example,  $x = 2$  corresponds to the results for systems ranked 2nd to 6th in human evaluation. A comparison is made between the ModernBERT$_{\text{Large}}$ without GED training and with additional training using binary-labeled GED.}
    \label{fig:window-analysis}
\end{figure}

Figure~\ref{fig:window-analysis} shows the results of a window analysis on system-level \texttt{SEEDA-S} using ModernBERT$_{\text{Large}}$, following the analysis by \citet{kobayashi-etal-2024-revisiting}. From Figure~\ref{fig:window-analysis}, we observe that additional training on the GED task improves the evaluation performance of top-ranked systems.

\begin{figure}[t]
    \centering
    \includegraphics[width=\linewidth]{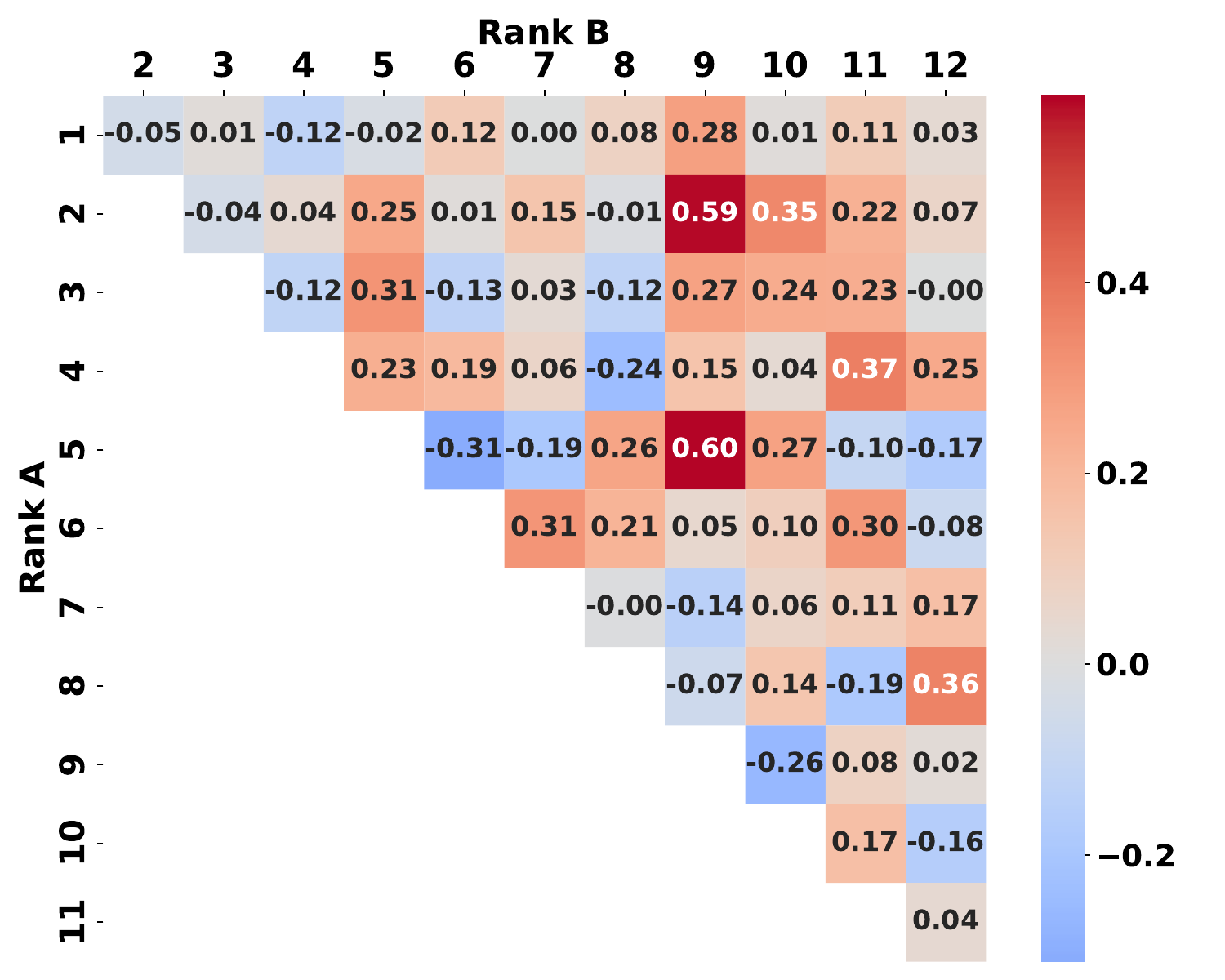}
    \caption{
    Results of pairwise sentence-level analysis.
    The hypothesis sentences for each source are ranked according to an evaluation metric, and all pairs are created from the hypotheses while retaining the rank information. Then, the same ranked pairs (A, B) are grouped together at the corpus level, and the percentage of agreement with the human evaluation is calculated for each pairs; $0 \leq A \leq 11$ and $A+1 \leq B \leq 12$, based on SEEDA's \texttt{Base} setting. The figure shows a heatmap of the differences in Kendall's tau after performing this pairwise analysis for each of the original IMPARA and IMPARA-GED. For example, for Rank A=2 and Rank B=9, the judgment in the hypothesis that the automatic evaluation ranked 2nd and 9th for each source. 
    The value of 0.59 at the cell indicates that the GED greatly improves the evaluation results in this pair.
    The difference is calculated between ModernBERT$_{\text{Large}}$ with and without additional training using binary-labeled GED.}
    \label{fig:pairwise-analysis}
\end{figure}

\paragraph{Sentence-level Pairwise Comparison.}
We investigate whether IMPARA-GED can distinguish the output quality of each GEC system in a pairwise comparison at the sentence level. Figure~\ref{fig:pairwise-analysis} shows the improvement in pairwise discriminative ability on \texttt{SEEDA-S} through the GED task. The results indicate that GED enhances the ability to distinguish between systems with more significant rank differences.

\section{Conclusion}
We proposed IMPARA-GED, a novel reference-free automatic GEC evaluation method, by enhancing the GED capabilities of PLMs. When using ModernBERT as the PLM, additional training on the binary-labeled GED task achieved the highest correlation with human evaluation among existing methods on SEEDA. Furthermore, window analysis revealed that IMPARA-GED improves evaluation performance, particularly for top-ranked systems.
Moreover, we revealed that current similarity estimators fail to adequately capture meaning preservation. We suggest that future development of GEC evaluation metrics may proceed in two directions: either by building a stronger similarity estimator or by enhancing the quality estimator. We publicly release IMPARA-GED as the official version, based on ModernBERT with binary classification, available at: \url{https://huggingface.co/naist-nlp/IMPARA-GED}.

\section{Limitations}

\paragraph{GED.}
We demonstrated that additional training with the GED task improves IMPARA’s performance. However, we did not determine which class type contributes the most to this improvement. Similarly, Yuan et al. and other studies related to GED-boosted GEC system construction have not explored which class type is most effective. Therefore, identifying the optimal number of class types is not the primary focus; rather, the key takeaway is that GED is effective for building automatic GEC evaluation models. That said, our findings suggest that even a two-class setting can yield improvements. Thus, determining the optimal class configuration is an important direction for future work.

\paragraph{Training method.}
We sequentially apply IMPARA after training on the GED task. This training approach allows for various future extensions, such as multi-task learning or adopting GRECO-style classifier types, e.g., word and gap labels~\cite{qorib-ng-2023-system}. The contribution of this paper lies in demonstrating that similarity filtering does not function effectively in reference-free GEC evaluation and that error-type classification, such as GED, has the potential to improve GEC evaluation. Therefore, we did not conduct a comprehensive investigation of alternative training strategies.
\paragraph{Parameter tuning.}
While we followed the training settings of \citet{maeda-etal-2022-impara} and \citet{yuan-etal-2021-multi}, further parameter tuning might lead to even better performance. However, to ensure a fair comparison with properly tuned results, we carefully monitored performance and conducted comparisons under a well-controlled experimental setup. Therefore, while we did not perform extensive parameter tuning, we believe that our evaluations and comparisons are sufficiently tuned to achieve the objectives of our study.%

\paragraph{Datasets.}
In this study, we used CoNLL-2013 and FCE to construct the models, following the setups of \citet{maeda-etal-2022-impara} and \citet{yuan-etal-2021-multi}. Therefore, leveraging high-quality, large-scale datasets such as W\&I+LOCNESS~\cite{bryant-etal-2019-bea} may lead to even better performance. Additionally, exploring data augmentation methods that enable effective GED training even with small-scale datasets like CoNLL-2013 remains a future challenge. However, in this study, we specifically investigated the impact of additional training with the GED task using the same dataset, ensuring robust and interpretable results. Therefore, while these points are important future directions, they are beyond the scope of this short paper.

\paragraph{Evaluation.}
In this study, we used SEEDA as the meta-evaluation dataset. SEEDA was introduced to address limitations in previous datasets, such as GJG~\cite{grundkiewicz-etal-2015-human}, which lacked coverage of modern neural network-based GEC models and suffered from a small number of system comparisons. SEEDA includes two benchmarks, \texttt{SEEDA-S} and \texttt{SEEDA-E}, both of which were used in our evaluation. Therefore, the evaluation conducted in this study is comprehensive and addresses current challenges in the field. Furthermore, while SEEDA is currently the only effective dataset available for automatic GEC evaluation, conducting evaluations on more specialized domains would be meaningful. However, due to resource limitations, this remains a broader challenge for the field rather than an issue specific to this study.

\paragraph{PLMs.}
IMPARA-GED can be applied to any encoder-based model. Therefore, leveraging other PLMs may enable the development of even higher-quality automatic GEC evaluation models. In this study, we aimed to verify the performance impact of additional training with the GED task. To this end, we conducted experiments using BERT$_{\text{Base}}$, which was originally used in IMPARA, as well as DeBERTa${_\text{Large}}$ and ModernBERT$_{\text{Large}}$, two of the latest improved versions of BERT. Furthermore, we conducted pilot studies with other PLMs and confirmed that additional training with the GED task consistently enhanced performance across different models. However, due to page limitations and resource constraints, we did not include these results in this paper. Thus, while our experiments fulfill the intended verification objective, achieving higher performance would require further evaluation with additional PLMs.

\section{Ethical Considerations}
In this study, we use open-source tools, PLMs, and datasets that are permitted for research use, ensuring no license issues. Additionally, all datasets used are publicly available and widely recognized in related research, guaranteeing that no harmful data was included in the experiments. For reproducibility, we provide the detailed settings in Appendices~\ref{sec:details-plms} and~\ref{sec:details-implementations}. Thus, this study has no ethical considerations.

\bibliography{anthology,custom}

\appendix

\section{Details of Each PLM}
\label{sec:details-plms}
We used the Hugging Face transformers library~\cite{wolf-etal-2020-transformers} for all experiments.
Table~\ref{tab:models} shows the PLMs we used in this study and their corresponding Hugging Face IDs.

\begin{table}[!th]
\centering
\resizebox{\linewidth}{!}{%
\setlength{\tabcolsep}{2pt}
\rowcolors{2}{gray!10}{white}
\begin{tabular}{@{}ll@{}}
\toprule
PLMs       &  HuggingFace ID                      \\
\midrule
BERT-Base           &  google-bert/bert-base-cased         \\ 
BERT-Large          &  google-bert/bert-large-cased        \\ 
BERT-Base-uncased   &  google-bert/bert-base-uncased       \\
BERT-Large-uncased  &  google-bert/bert-large-uncased      \\ 
ELECTRA-Base        &  google/electra-base-discriminator   \\ 
ELECTRA-Large       &  google/electra-large-discriminator  \\ 
DeBERTa-v3-Base     &  microsoft/deberta-v3-base          \\ 
DeBERTa-v3-Large    &  microsoft/deberta-v3-large          \\ 
ModernBERT-Base     &  answerdotai/ModernBERT-base         \\ 
ModernBERT-Large    &  answerdotai/ModernBERT-large        \\ 
\bottomrule
\end{tabular}
}
\caption{Lists of the PLMs we used in this study and their corresponding Hugging Face IDs.}
\label{tab:models}
\end{table}

\section{Details of Implementations}
\label{sec:details-implementations}
For the training setup of IMPARA-GED, we followed the settings of \citet{yuan-etal-2021-multi} for building the GED model and \citet{maeda-etal-2022-impara} for constructing the quality evaluation model.
For GED training, we used \texttt{ged\_baselines}\footnote{\url{https://github.com/gotutiyan/ged_baselines}}.
For IMPARA training, we used its public reproduction implementation\footnote{\url{https://github.com/gotutiyan/IMPARA}}.
The quality estimator, IMPARA-QE, we used in Table~\ref{tab:impara-se}, a public reproduction model available on Hugging Face\footnote{\url{https://huggingface.co/gotutiyan/IMPARA-QE}}.
Unless otherwise specified, we used the default hyperparameters of these tools.
We used a single NVIDIA GeForce RTX 3090 GPU for all experiments.
We are ready to publish all experimental codes after acceptance to ensure reproducibility. Additionally, IMPARA-GED weights are made publicly available\footnote{\url{https://huggingface.co/naist-nlp/IMPARA-GED}}.

\end{document}